\documentclass[11pt]{article}

\usepackage{acl}
\usepackage{times}
\usepackage{latexsym}

\usepackage[T1]{fontenc}
\usepackage[utf8]{inputenc}

\usepackage{microtype}
\usepackage{amsmath, amssymb}
\usepackage{booktabs}
\usepackage{graphicx}
\usepackage{multirow}
\usepackage{subfigure}
\usepackage{algorithm}
\usepackage{algorithmic}
\usepackage{bbm}
\usepackage{subfigure}

\newcommand\blfootnote[1]{%
\begingroup
\renewcommand\thefootnote{}\footnote{#1}%
\addtocounter{footnote}{-1}%
\endgroup
}

\title{Pruning Pre-trained Language Models Without Fine-Tuning}

 \author{
     Ting Jiang\textsuperscript{\rm 1}, Deqing Wang\textsuperscript{\rm 13}$^\dagger$, Fuzhen Zhuang\textsuperscript{\rm 123}, Ruobing Xie\textsuperscript{\rm 4}, Feng Xia\textsuperscript{\rm 4}\\
     \textsuperscript{\rm 1}SKLSDE Lab, School of Computer, Beihang University, Beijing, China \\
     \textsuperscript{\rm 2}Institute of Artificial Intelligence, Beihang University, Beijing, China\\
     \textsuperscript{\rm 3} Zhongguancun Laboratory, Beijing, China \textsuperscript{\rm 4} WeChat, Tencent, Beijing, China \\
     \{royokong, dqwang, zhuangfuzhen\}@buaa.edu.cn
 }

\begin{document}
\maketitle

\begin{abstract}

To overcome the overparameterized problem in Pre-trained Language Models (PLMs),
pruning is widely used as a simple and straightforward compression method by directly removing unimportant weights.
Previous first-order methods successfully compress PLMs to extremely high sparsity with little performance drop.
These methods, such as movement pruning, use first-order information to prune PLMs while fine-tuning the remaining weights.
%Even though we find that first-order pruning is able to converge PLMs to downstream tasks without fine-tuning, these methods still fine-tune the remaining weights during pruning.
%These methods successfully compress PLMs to extremely high sparsity with little performance drop.
%while keeping accuracy loss to a small degree.
%However, even with more training parameters, these first-order pruning methods still underperform zero-order at low sparsity.
%In this work, we find first-order information is underestimated for pruning PLMs.
%even first-order pruning is able to converge PLMs to downstream tasks without fine-tuning.
In this work,
we argue fine-tuning is redundant for first-order pruning, since first-order pruning is sufficient to converge PLMs to downstream tasks without fine-tuning.
%we find first-order pruning is sufficient to converge PLMs to downstream tasks without fine-tuning.
%It limits their performance.
Under this motivation,
%To this end, we propose a first-order pruning method without fine-tuning.
we propose Static Model Pruning (SMP), which only uses first-order pruning to adapt PLMs to downstream tasks while achieving the target sparsity level.
In addition, we also design a new masking function and training objective to further improve SMP.
%Instead of pruning and fine-tuning PLMs in previous methods, our method freezes PLMs and relies on pruning to replace fine-tuning.
Extensive experiments at various sparsity levels show SMP has significant improvements over first-order and zero-order methods.%, which demonstrates the redundancy of fine-tuning in first order pruning methods.
Unlike previous first-order methods, SMP is also applicable to low sparsity and outperforms zero-order methods.
%Even under low sparsity, preivous first-order methods find them underperform zero-order methods at low sparsity, SMP still outperforms all other methods.
Meanwhile, SMP is more parameter efficient than other methods due to it does not require fine-tuning. Our code is available at \url{https://github.com/kongds/SMP}.

%Magnitude pruning is widely used as a simple zero-order pruning method.
%Movement pruning
\end{abstract}
 \blfootnote{ $\dagger$ Corresponding Author.}

\section{Introduction}
Pre-trained Language Models (PLMs) like BERT~\cite{devlin-etal-2019-bert} have shown powerful performance in natural language processing by transferring the knowledge from large-scale corpus to downstream tasks.
These models also require large-scale parameters to cope with the large-scale corpus in pretraining.
However, these large-scale parameters are overwhelming for most downstream tasks~\cite{Chen2020e}, which results in significant overhead for transferring and storing them.

To compress PLM, pruning is widely used by removing unimportant weights and setting them to zeros.
By using sparse subnetworks instead of the original complete network, existing pruning methods can maintain the original accuracy by removing most weights.
%For example, movement pruning reaches 95\% of the original performance with only 3\% of the model parameters~\cite{Sanh2020}.
 Magnitude pruning~\cite{magnitude} as a common method uses zeroth-order information to make pruning decisions based on the absolute value of weights.
%Magnitude pruning is widely used in the Lottery Ticket Hypothesis (LTH)~\cite{frankle2018lottery}.
%Based on this hypothesis, current works~\cite{Chen2020e, Liang2021} not only find existing subnetworks achieve the original performance but also exceed.
However, in the process of adapting to downstream tasks, the weight values in PLMs are already predetermined from the original values.
To overcome this shortcoming, movement pruning~\cite{Sanh2020} uses first-order information to select weights based on how they change in training rather than their absolute value.
%Although movement pruning strongly outperforms magnitude pruning at high sparsity, magnitude pruning still achieves better performance at low sparsity~\cite{Sanh2020}.
To adapt PLMs for downstream tasks,
most methods like movement pruning perform pruning and fine-tuning together by gradually increasing the sparsity during training.
With the development of the Lottery Ticket Hypothesis (LTH) ~\cite{frankle2018lottery} in PLMs,
some methods~\cite{Chen2020e, Liang2021} find certain subnetworks from the PLM by pruning, and then fine-tune these subnetworks from pre-trained weights.
Moreover, if the fine-tuned subnetwok can match the performance of the full PLM, this subnetwork is called winning ticket~\cite{Chen2020e}.

In this work, we propose a simple but efficient first-order method.
Contrary to the previous pruning method, our method adapts PLMs by only pruning, without fine-tuning.
It makes pruning decisions based on the movement trend of weights, rather than actual movement in movement pruning.
To improve the performance of our method, we propose a new masking function to better align the remaining weights according to the architecture of PLMs.
We also avoid fine-tuning weights in the task-specific head by using our head initialization method.
By keeping the PLM frozen, we can save half of the trainable parameters compared to other first-order methods, and only introduce a binary mask as the new parameter for each downstream task at various sparsity levels.
%Our method is also efficient in parameters for new tasks.
%Instead of saving the parameters of the subnetwork,  we only use a binary mask to represent the remaining weights in PLMs.
%Our method strongly outperforms the current state-of-the-art pruning methods from low sparsity to high.
%By leveraging our method, we also find there are not only winning tickets in PLMs but some subnetworks that can match the original performance by only pruning based on first-order information.
Extensive experiments on a wide variety of sparsity demonstrate our methods strongly outperform state-of-the-art pruning methods.
Contrary to previous first-order methods~\cite{Sanh2020}, which show poor performance at low sparsity,
our method is also applied to low sparsity and achieves better performances than zero-order methods.
%Previous work~\cite{Sanh2020} find first-order pruning methods show worse performance than zero-order at low sparsity~\cite{Sanh2020}.
%Our first-order pruning method can strongly

%Based on our method, we
%Contrary to the previous pruning methods in PLM, our method only prune PLMs without fine-tuning the subnetwork.
%A sparsity network  whole network.
%These methods can be divided into two steps: 1) finding subnetworks, and 2) fine-tuning subnetworks.
%One can perform both steps simultaneously, such as movement pruning, or sequentially, such as iterative magnitude pruning with rewinding.
%Previous works~\cite{Chen2020e, Liang2021} show the importance of finding a suitable subnetwork for downstream tasks.
%Moreover, \citeauthor{Liang2021} even finds these subnetworks outperform the full model.

\section{Related Work}

%\subsection{Pruning in Pre-trained Language Models}

Compressing PLMs for transfer learning is a popular area of research.
Many compression methods are proposed to solve overparameterized problem in PLMs, such as model pruning~\cite{magnitude, taylor, xia2022structured}, knowledge distillation~\cite{tinybert, minilm}, quantization~\cite{qbert, Qin2022}, and matrix decomposition~\cite{albert}.
Among them, pruning methods have been widely studied as the most intuitive approach.

Pruning methods focus on identifying and removing unimportant weights from the model.
%Current works~\cite{Sanh2020, Xu2021d} have shown PLMs can compressed by removing 97\% of the weights with little performance dropping.
%Based on the way of identifying important weights, current pruning methods based on Taylor expansion can be roughly divided into zero-order methods, first-order methods.
%Based on the way of identifying important weights,
Zero-order methods and first-order methods are widely used to prune PLMs.
%current pruning methods based on Taylor expansion can be roughly divided into zero-order methods, first-order methods.
For zero-order methods, magnitude pruning~\cite{magnitude} simply prunes based on absolute value of their weights. %, which is a simple and effective pruning method.
For first-order methods, which are based on first-order Taylor expansion to make pruning decision,
$L_0$ regularization~\cite{louizos2017learning} adds the $L_0$ norm regularization to decrease remaining weights by sampling them with hard-concrete distribution.
Movement pruning~\cite{Sanh2020} uses \textit{straight-through estimator}~\cite{bengio2013estimating} to calculate first-order informantion.

%\subsection{Lottery Ticket Hypothesis}
Based on pruning methods, \citeauthor{frankle2018lottery}~(\citeyear{frankle2018lottery}) proposes Lottery Ticket Hypothesis (LTH).
LTH clarifies the existence of sparse subnetworks (i.e., winning tickets) that can achieve almost the same performance as the full model when trained individually.
With the development of LTH, lots of works that focus on the PLMs have emerged.
\citeauthor{Chen2020e}~(\citeyear{Chen2020e}) find that BERT contains winning tickets with a sparsity of 40\% to 90\%, and the winning ticket in the mask language modeling task can be transferred to other downstream tasks.
Recent works also try to leverage LTH to improve the performance and efficiency of PLM.
\citeauthor{Liang2021}~(\citeyear{Liang2021}) find generalization performance of the winning tickets first improves and then deteriorates after a certain threshold. By leveraging this phenomenon, they show LTH can successfully improve the performance of downstream tasks.
%\citeauthor{Zheng2022a} propose a general efficient training algorithm applicable to both pretraining and fine-tuning of PLM based on LTH.

\section{Background}
Let $\mathbf{a}=\mathbf{Wx}$ refer to a fully-connected layer in PLMs, where $\mathbf{W}\in\mathbb{R}^{n \times n}$ is the weight matrix, $\mathbf{x}\in\mathbb{R}^n$ and $\mathbf{a} \in \mathbb{R}^n$ are the input and output respectively.
The pruning can be represented by $\bf a = (W \odot M)x$, where $\mathbf{M} \in \{0, 1\}^{n \times n}$ is the  binary mask.

We first review two common pruning methods in PLMs: magnitude pruning~\citep{magnitude} and movement pruning~\cite{Sanh2020}.
Magnitude pruning relies on the zeroth-order information to decide $\bf M$ by keeping the top $v$ percent of weights according to their absolute value  $\mathbf{M}=\text{Top}_v(\mathbf{S})$. The importance scores $\mathbf{S} \in \mathbb{R}^{n \times n}$ is:
\begin{equation}
\begin{split}
S_{i,j}^{(T)}&= \left|W_{i, j}^{(T)}\right|\\
&= \left|W_{i,j} - \alpha_w\sum_{t < T}\left(\frac{\partial \mathcal{L}}{\partial W_{i, j}}\right)^{(t)}\right|
\end{split}
\end{equation}
where $S_{i,j}^{(T)}$ is the importance score corresponding to $W_{i,j}^{(T)}$ after $T$ steps update, $\mathcal{L}$ and $\alpha_w$ are learning objective and learning rate of $W_{i,j}$. %Compared to magnitude pruning, movement pruning selects weights that are increasing their absolute value.J
Magnitude pruning selects weights with high absolute values during fine-tuning.

For movement pruning, it relies on the first-order information by learning the importance scores $\mathbf{S}$ with gradient.
The gradient of $\mathbf{S}$ is approximated with the \textit{staight-through estimator}~\cite{bengio2013estimating}, which directly uses the gradient from $\bf M$.
According to~\cite{Sanh2020}, the importance scores $\mathbf{S}$ is:
\begin{equation}
S_{i,j}^{(T)} = - \alpha_s\sum_{t < T}\left(\frac{\partial \mathcal{L}}{\partial W_{i, j}}\right)^{(t)} W_{i, j}^{(t)}
\end{equation}
where $\alpha_s$ is the learning rate of $\mathbf{S}$.
Compared to magnitude pruning, movement pruning selects weights that are increasing their absolute value.

%Both pruning methods prune the model during training.
To achieve target sparsity,  one common method is \textit{automated gradual pruning}~\cite{cube}.
The sparsity level $v$ is gradually increased with a cubic sparsity scheduler starting from the training step $t_0$: $v^{t}=v_{f}+\left(v_{0}-v_{f}\right)\left(1-\frac{t-t_{0}}{N \Delta t}\right)^{3}$, where  $v_{0}$ and $v_{f}$ are the initial and target sparsity, $N$ is overall pruning steps, and $\Delta t$ is the pruning frequency.

During training, these methods update both $\mathbf{W}$ and $\mathbf{S}$ to perform pruning and fine-tuning simultaneously.
Since fine-tuned weights stay close to their pre-trained values~\cite{Sanh2020},
%magnitude pruning underperfoms movement pruning due to it determine importance socres according to absolute values.
the importance scores of magnitude pruning is influenced by pre-trained values, which limits its performance at high sparsity. However, magnitude pruning still outperforms movement pruning at low sparsity.

%\section{Methodology}
\section{Static Model Pruning}
In this work,
we propose a simple first-order pruning method called Static Model Pruning (SMP).
It freezes $\mathbf{W}$  to make pruning on PLMs more efficient and transferable.
Based on movement pruning~\cite{Sanh2020},
our importance scores $\mathbf{S}$ is:
\begin{equation}
\label{eq:score}
S_{i,j}^{(T)} = - \alpha_sW_{i, j}\sum_{t < T}\left(\frac{\partial \mathcal{L}}{\partial W^\prime_{i, j}}\right)^{(t)}
\end{equation}
where $W^\prime_{i,j}$ is $W_{i,j}M_{i,j}$.
Since our method freezes $W_{i,j}$, we also keep the binary masking term $M_{i,j}$.
$S_{i,j}$ is increasing when $W_{i,j}\frac{\partial \mathcal{L}}{\partial W^\prime_{i, j}} < 0$.
For remaining weight $W^\prime_{i,j}=W_{i,j}$,
it means that movement trending $-\frac{\partial \mathcal{L}}{\partial W^\prime_{i, j}}$ increases the absolute value of $W_{i,j}$.
%$S_{i,j}$ increases when $W_{i,j}\frac{\partial \mathcal{L}}{\partial W_{i, j}} < 0$ ~\cite{Sanh2020}.
%More specificaly,
%the movement trending $-\frac{\partial \mathcal{L}}{\partial W_{i, j}}$ that increases the absolute value of $W_{i,j}$  will lead to $S_{i,j}$ increase.
For removed weight $W^\prime_{i,j}=0$, %$S_{i,j}$ increases when the movement trending of $W^\prime_{i,j}$
it means that movement trending encourages $0$ to close $W_{i,j}$.
%since $W_{i,j}$ is replaced by 0, the movement trending is corespondent to 0

%More specifically, when $W^_{i,j}$ is not pruned ( $M_{i,j} = 1$ ),
%$S_{i,j}$ is increasing or decreasing according to the trending of $W_{i,j}$

\subsection{Masking Function}
To get masks $\mathbf{M}$ based on $\mathbf{S}$, we consider two masking functions according to the pruning structure: local and global.

For the local masking function, we simply apply the Top$_v$ function to each matrix: $\mathbf{M}=\text{Top}_v(\mathbf{S})$, which selects the $v\%$ most importance weights according to $\mathbf{S}$ matrix by matrix.%, following in previous pruning methods~\cite{Sanh2020}.

For the global masking function, ranking all importance scores together (around 85M in BERT base) is computationally inefficient,
which even harms the final performance in section~\ref{sec:masking}.
%which even harms the final performance compared to local masks in our settings.
To this end, we propose a new global masking function that assigns sparsity levels based on the overall score of each weight matrix.
Considering the architecture of BERT, which has $L$ transformer layers, each layer contains a self-attention layer and a feed-forward layer.
In $l$th self-attention block, $\mathbf{W}_Q^l$, $\mathbf{W}_K^l$, $\mathbf{W}_V^l$, and $\mathbf{W}_O^l$ are the weight matrices we need to prune.
In the same way, $\mathbf{W}_U^l$ and $\mathbf{W}_D^l$ are the matrices to be pruned in the $l$th feed-forward layer.
%Contrary to the previous method,
We first calculate the sparsity level of each weight matrix instead of ranking all parameters of the network.
The sparsity level of each weight matrix $v_{(\cdot)}^l$ is computed as follows:
\begin{equation}
\label{eq:our_v}
v_{(\cdot)}^l = \frac{R\left(\mathbf{S}_{(\cdot)}^l\right)L}{\sum_{l^\prime=1}^{L}R\left(\mathbf{S}_{(\cdot)}^{l^\prime}\right)} v
\end{equation}
where $R(\mathbf{S})=\sum_{i,j}\sigma(S_{i,j})$ is the regularization term of $\mathbf{S}$ with sigmoid $\sigma$, $\mathbf{S}_{(\cdot)}^{l}$ is the importance socres of weight $\mathbf{W}_{(\cdot)}^{l}$, and $(\cdot)$ can be one of $\{ Q, K, V, O, U, D\}$.
The sparsity level is determined by the proportion of important scores to the same type of matrix in different layers.

\subsection{Task-Specific Head}
Instead of training the task-specific head from scratch,
we initialize it from BERT token embedding and keep it frozen during training.
Inspired by current prompt tuning methods, we initialize the task-specific head according to BERT token embeddings of corresponding label words following \cite{gao2020making}.
For example, we use token embeddings of ``great'' and ``terrible'' to initialize classification head in SST2, and the  predicted positive label score is $h_{\texttt{[CLS]}}e_{\texttt{great}}^T$, where $h_{\texttt{[CLS]}}$ is the final hidden state of the special token \texttt{[CLS]} and $e_{\texttt{great}}$ is the token embeddings of ``great''.

\subsection{Training Objective} \label{sec:to}
To prune the model, we use the cubic sparsity scheduling~\cite{cube} without warmup steps. The sparsity $v_{t}$ at $t$ steps is:
\begin{equation}
v_{t}=\begin{cases} v_{f}-v_{f}\left(1-\frac{t}{N}\right)^{3} & t<N \\ v_{f} & \text { o.w. }\end{cases}
\end{equation}
we gradually increase sparsity from 0 to target sparsity $v_f$ in the first $N$ steps.
After $N$ steps, we keep the sparsity $v_{t}=v_f$.
During this stage, the number of remaining weights remains the same,
but these weights can also be replaced with the removed weights according to important scores.

%constant and allow masks to be swapped between the removed weights and the remaining weights.

We evaluate our method with and without knowledge distillation.
For the settings without knowledge distillation, we optimize the following loss function:
\begin{equation}
  \mathcal{L}=\mathcal{L}_{\mathrm{CE}}+\lambda_R\frac{v_{t}}{v_{f}} R\left(\mathbf{S}\right)
\end{equation}
where $\mathcal{L}_{\mathrm{CE}}$ is the classification loss corresponding to the task and $R\left(\mathbf{S}\right)$
is the regularization term with hyperparameter $\lambda_R$.
Inspired by soft-movement~\cite{Sanh2020}, it uses a regularization term to decrease $\mathbf{S}$ to increase sparsity with the thresholding masking function.%$\sigma\left(\mathbf{S}\right)>\tau$.
%We find the regularization term is also important for the masking function like Top$_v$.
We find the regularization term is also important in our method.
%Since $\mathbf{S}$ is initialized by 0 and $\lambda_R$ is large enough,
Since $\lambda_R$ is large enough in our method,
the most important scores in $\mathbf{S}$ are less than zero when the current sparsity level $v_t$ is close to $v_f$.
Due to the gradient $\frac{\partial R(\mathbf{S})}{\partial S_{i,j}}=\frac{\partial \sigma(S_{i,j})}{\partial S_{i,j}}$ increases with the increase of $S_{i,j}$ when $S_{i,j} < 0$,
scores corresponding to the remaining weights will have a larger penalty than removed weights.
It encourages the $\mathbf{M}$ to be changed when $v_t$ is almost reached or reached $v_f$.

For the settings with knowledge distillation,
we simply add a distillation loss $\mathcal{L}_\mathrm{KD}$ in $\mathcal{L}$ following \cite{Sanh2020, xu2021dense}:
\begin{equation}
  \mathcal{L}_\mathrm{KD}=D_{\mathrm{KL}}\left(\mathbf{p}_{s} \| \mathbf{p}_{t}\right)
\end{equation}
where $D_\mathrm{KL}$ is the KL-divergence. $\mathbf{p}_{s}$ and $\mathbf{p}_{t}$ are output distributions of the student model and teacher model.

\begin{table*}[h]
\centering
\resizebox{0.9\textwidth}{!}{%
    \begin{tabular}{lccccccc}
    \toprule
    %\bf Methods & \bf Sparsity & \bf Trainable Parameters & \bf MNLI-m/-mm & \bf QQP$_{\mathrm{ACC/F1}}$ & \bf SQuAD$_{\mathrm{EM/F1}}$ \\
    %\bf Methods & \bf Remaining & \bf New Params & \bf Trainable & \bf MNLI & \bf QQP & \bf SQuAD \\
    %\bf & \bf Weights & \bf Per Task& \bf Params & \bf $_{\mathrm{ACC/MM_{ACC}}}$  & \bf $_{\mathrm{ACC/F1}}$ & \bf $_{\mathrm{EM/F1}}$ \\
     Methods &  Remaining &  New Params &  Trainable &  MNLI &  QQP &  SQuAD \\
     &  Weights &  Per Task&  Params &  $_{\mathrm{M_{ACC}/MM_{ACC}}}$  &  $_{\mathrm{ACC/F1}}$ &  $_{\mathrm{EM/F1}}$ \\
    \midrule
    BERT$_{\mathrm{base}}$ &  100\% & 110M & 110M & 84.5/84.9 & 91.4/88.4  & 80.4/88.1 \\
    \midrule
    \midrule
    \multicolumn{7}{c}{\emph{Without Knowledge Distillation}} \\
    \midrule
    \midrule
    Movement~\citep{Sanh2020} & 10\% & 8.5M + $\theta_\textbf{M}$ & 170M & 79.3/79.5 & 89.1/85.5 & 71.9/81.7 \\
    Soft-Movement~\citep{Sanh2020} & 10\% & 8.5M + $\theta_\textbf{M}$ & 170M &  80.7/81.1 & 90.5/87.1 & 71.3/81.5\\
    SMP-L (Our) & 10\% & $\theta_\textbf{M}$ & 85M & 82.0/82.3 & \bf 90.8/87.7 & 75.0/84.3 \\
    SMP-S (Our) & 10\% & $\theta_\textbf{M}$ & 85M& \bf82.5/82.3 & 90.8/87.6 & \bf 75.1/84.6 \\
    \midrule
    Movement~\citep{Sanh2020} & 3\% & 2.6M+$\theta_\textbf{M}$ & 170M & 76.1/76.7 & 85.6/81.0 & 65.2/76.3 \\
    Soft-Movement~\citep{Sanh2020} & 3\% & 2.6M+$\theta_\textbf{M}$ & 170M & 79.0/79.6 & 89.3/85.6 & 69.5/79.9 \\
    SMP-L (Our) & 3\% & $\theta_\textbf{M}$ & 85M & 80.6/81.0 & 90.2/87.0 & 70.7/81.0 \\
    SMP-S (Our) & 3\% & $\theta_\textbf{M}$ & 85M& \bf 80.9/81.1 & \bf 90.3/87.1 & \bf 70.9/81.4  \\
    \bottomrule
    \midrule

    \multicolumn{7}{c}{\emph{With Knowledge Distillation}} \\
    \midrule
    \midrule
    Movement~\citep{Sanh2020}   & 50\% &  42.5M+$\theta_\textbf{M}$ & 170M & 82.5/82.9 & 91.0/87.8 & 79.8/87.6 \\
    CAP~\citep{xu2021dense} & 50\% & 42.5M+$\theta_\textbf{M}$ & 170M &  83.8/84.2 & 91.6/88.6 &  80.9/88.2 \\
    SMP-L (Our) & 50\% &$\theta_\textbf{M}$& 85M& 85.3/\textbf{85.6} &   91.6/88.7 &  82.2/89.4 \\
    SMP-S (Our) & 50\% &$\theta_\textbf{M}$& 85M& \textbf{85.7}/85.5 &  \bf 91.7/88.8 & \bf 82.8/89.8 \\
    \midrule
Magnitude~\citep{magnitude} & 10\% &8.5M+$\theta_\textbf{M}$& 85M & 78.3/79.3 & 79.8/75.9 & 70.2/80.1 \\
    L$_0$-regularization~\citep{l0} & 10\% &8.5M+$\theta_\textbf{M}$& 170M & 78.7/79.7 & 88.1/82.8 & 72.4/81.9 \\
    Movement~\citep{Sanh2020} & 10\% &8.5M+$\theta_\textbf{M}$& 170M & 80.1/80.4 & 89.7/86.2 & 75.6/84.3 \\
    Soft-Movement~\citep{Sanh2020} & 10\% &8.5M+$\theta_\textbf{M}$& 170M & 81.2/81.8 & 90.2/86.8 & 76.6/84.9 \\
    %CAP-m~\cite{xu2021dense} & 10\% &8.5M+$\theta_\textbf{M}$& 170M & 81.1/81.8 & 90.5/87.2 & 76.5/85.1 \\
    CAP~\cite{xu2021dense} & 10\% &8.5M+$\theta_\textbf{M}$& 170M &  82.0/82.9 &  90.7/87.4 &  77.1/85.6 \\
    SMP-L (Our) & 10\% &$\theta_\textbf{M}$& 85M& 83.1/83.1 & 91.0/87.9 & 78.9/86.9 \\
    SMP-S (Our) & 10\% &$\theta_\textbf{M}$& 85M& \bf 83.7/83.6 &  \bf 91.0/87.9 & \bf 79.3/87.2 \\
    \midrule
    Movement~\citep{Sanh2020} & 3\% &  2.6M+$\theta_\textbf{M}$ & 170M & 76.5/77.4 & 86.1/81.5 & 67.5/78.0 \\
    Soft-Movement~\citep{Sanh2020} & 3\% &  2.6M+$\theta_\textbf{M}$ & 170M & 79.5/80.1 & 89.1/85.5 & 72.7/82.3 \\
    %CAP-m~\citep{xu2021dense} & 3\% &  2.6M+$\theta_\textbf{M}$ & 170M & 77.5/78.4 & 88.8/85.0 & 69.5/79.7 \\
    CAP~\cite{xu2021dense} & 3\% &  2.6M+$\theta_\textbf{M}$ & 170M &  80.1/81.3 &  90.2/86.7 &  73.8/83.0 \\
    SMP-L (Our) & 3\% &$\theta_\textbf{M}$& 85M& 80.8/81.2 & 90.1/87.0 & 74.0/83.4\\
    SMP-S (Our) & 3\% &$\theta_\textbf{M}$& 85M& \bf 81.8/82.0 & \bf90.5/87.4 & \bf75.0/84.1\\
    \bottomrule
    \end{tabular}
    }
\caption[]{
  Performance at high sparsity.
    %``Trainable params''
  %``New params per task'' is the new parameters added to a task based on BERT.
  SMP-L and SMP-S refer to our method with local masking function and our masking function.
  $\theta_\textbf{M}$ is the size of binary mask $\textbf{M}$, which is around 2.7M parameters and can be further compressed.\footnotemark
  Since other pruning methods freeze the embedding modules of BERT~\cite{Sanh2020}, the trainable parameters of first-order methods are the sum of BERT encoder (85M), importance scores $\textbf{S}$ (85M) and task-specific head (less than 0.01M).
  For zero-order pruning methods like magnitude pruning, the trainable parameters are 85M, excluding $\textbf{S}$.
  Our results are averaged from five random seeds.
}
\label{table:main}
\end{table*}

\begin{figure*}[h]
\centering    %居中

\subfigure[MNLI]
{
	\includegraphics[width=0.66\columnwidth]{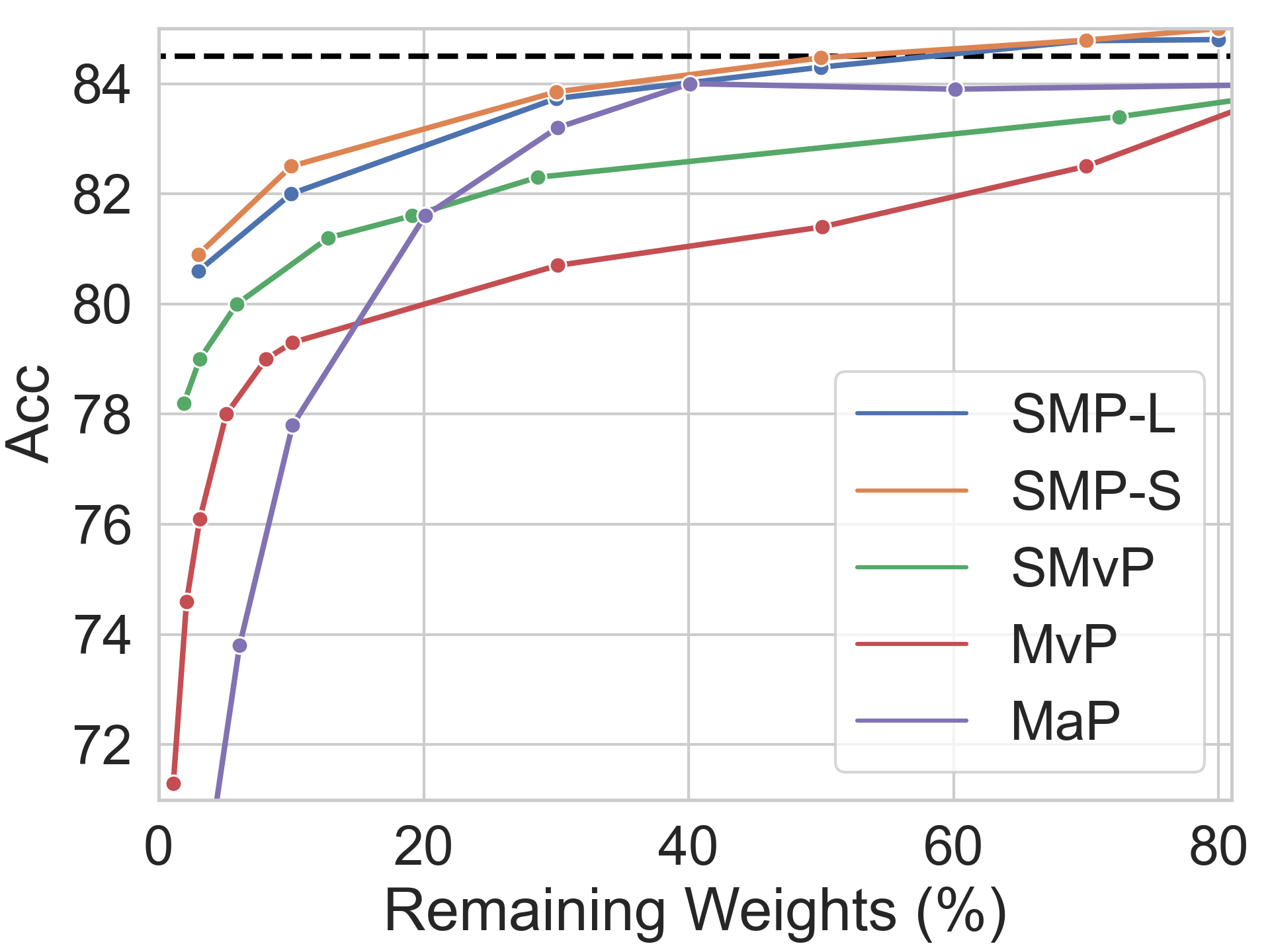}
}
\subfigure[QQP]
{
	\includegraphics[width=0.66\columnwidth]{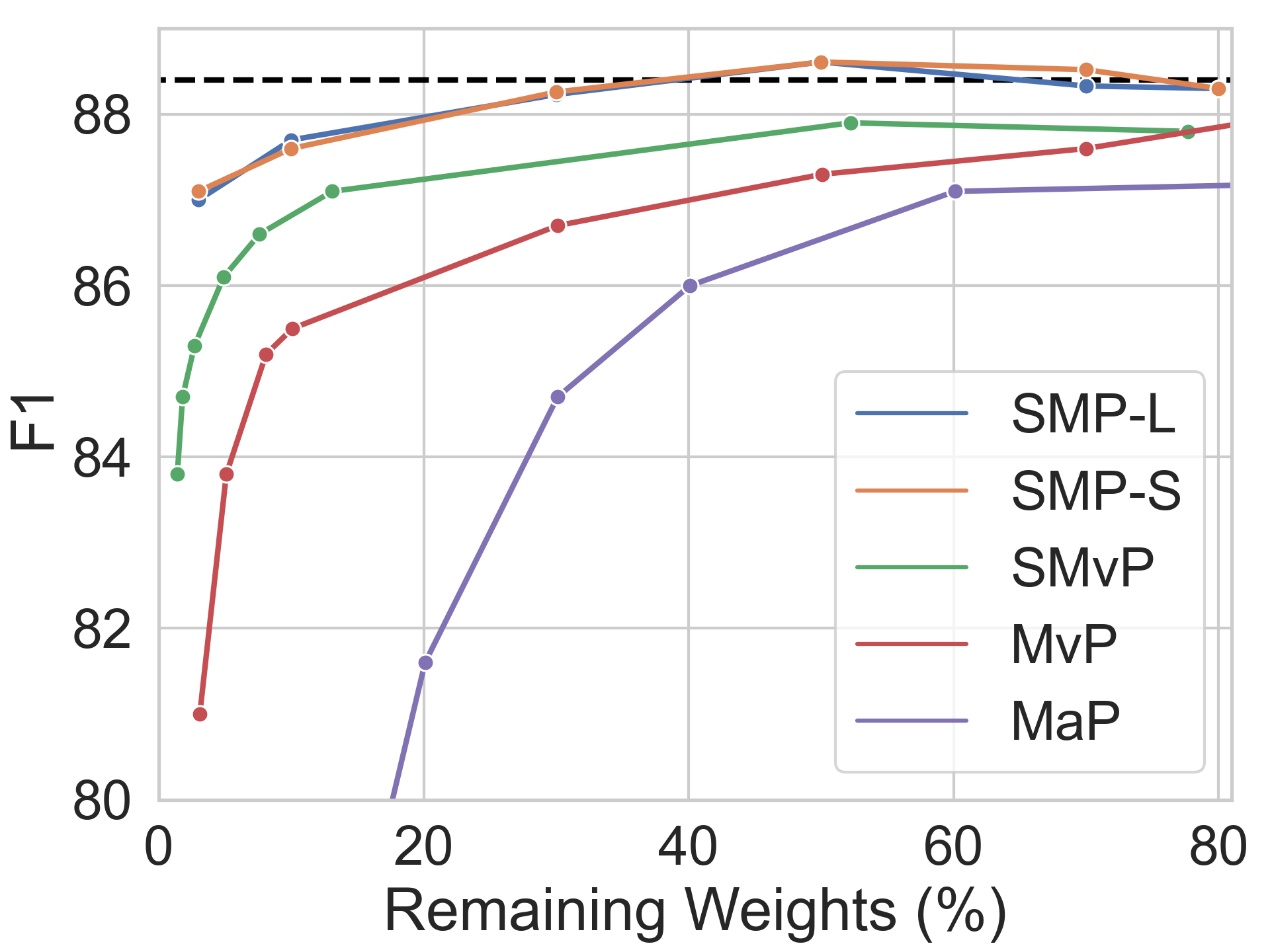}
}
\subfigure[SQuAD]
{
	\includegraphics[width=0.66\columnwidth]{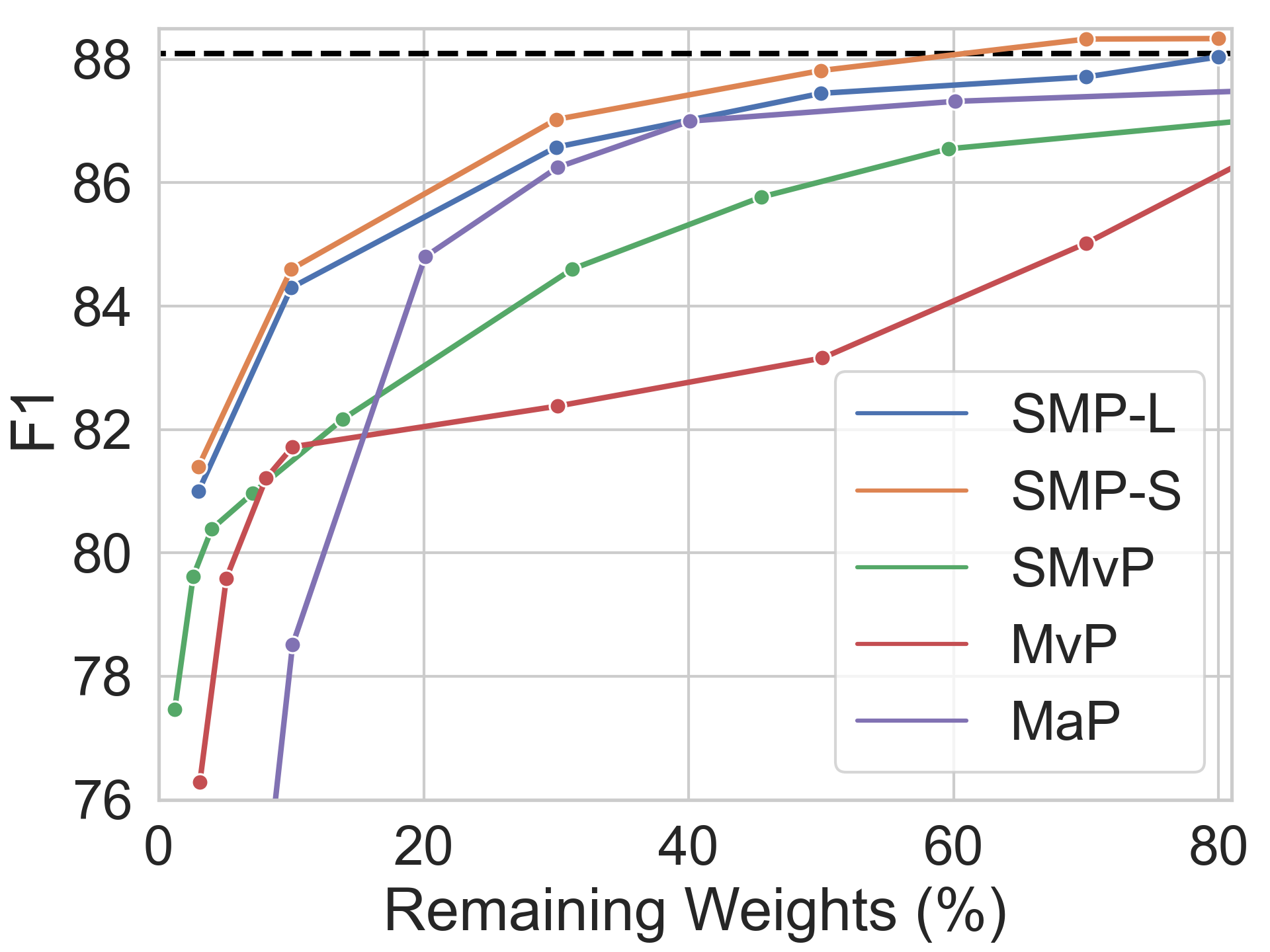}
}

\subfigure[MNLI + KD]
{
	\includegraphics[width=0.66\columnwidth]{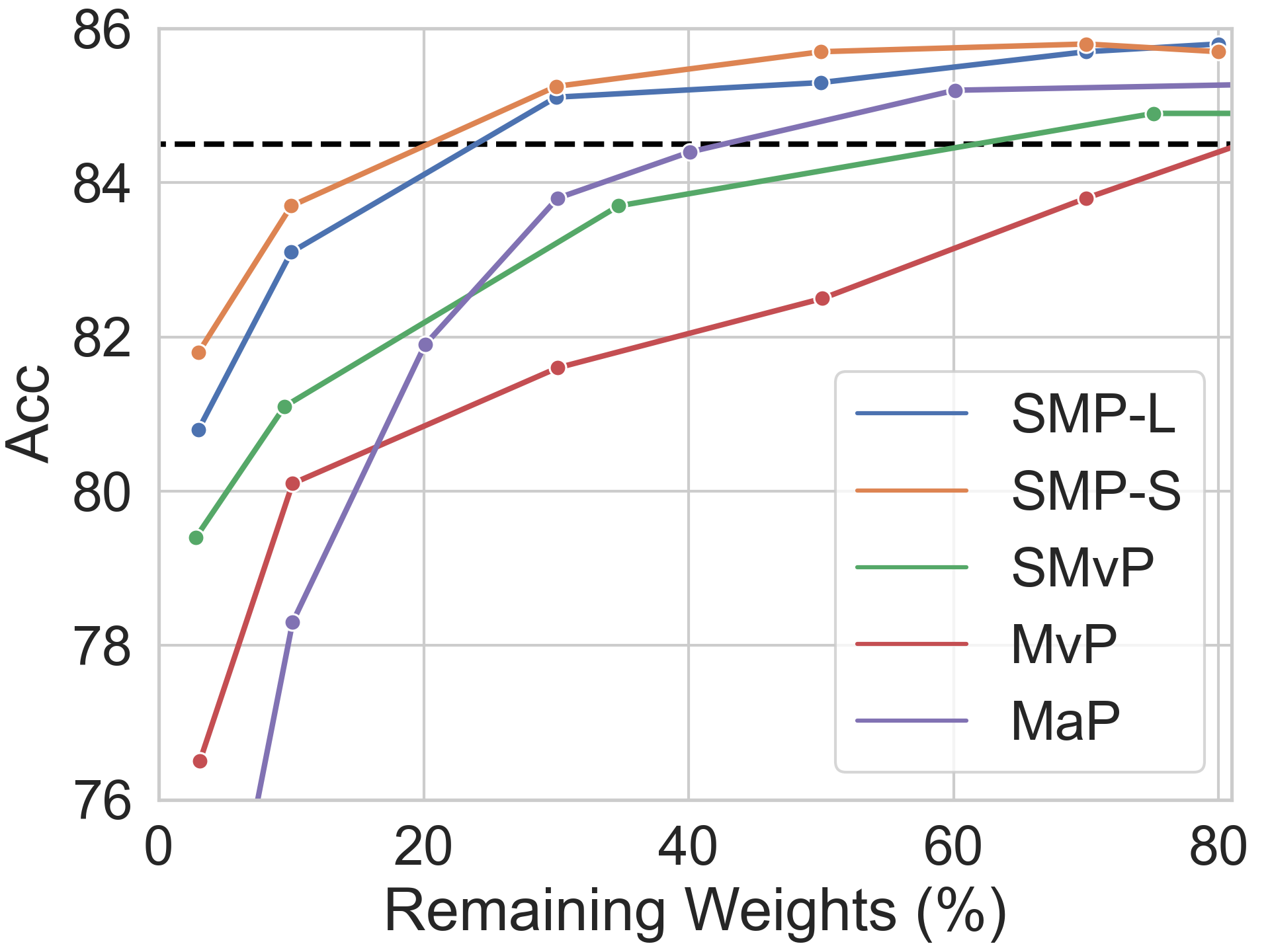}
}
\subfigure[QQP + KD]
{
	\includegraphics[width=0.66\columnwidth]{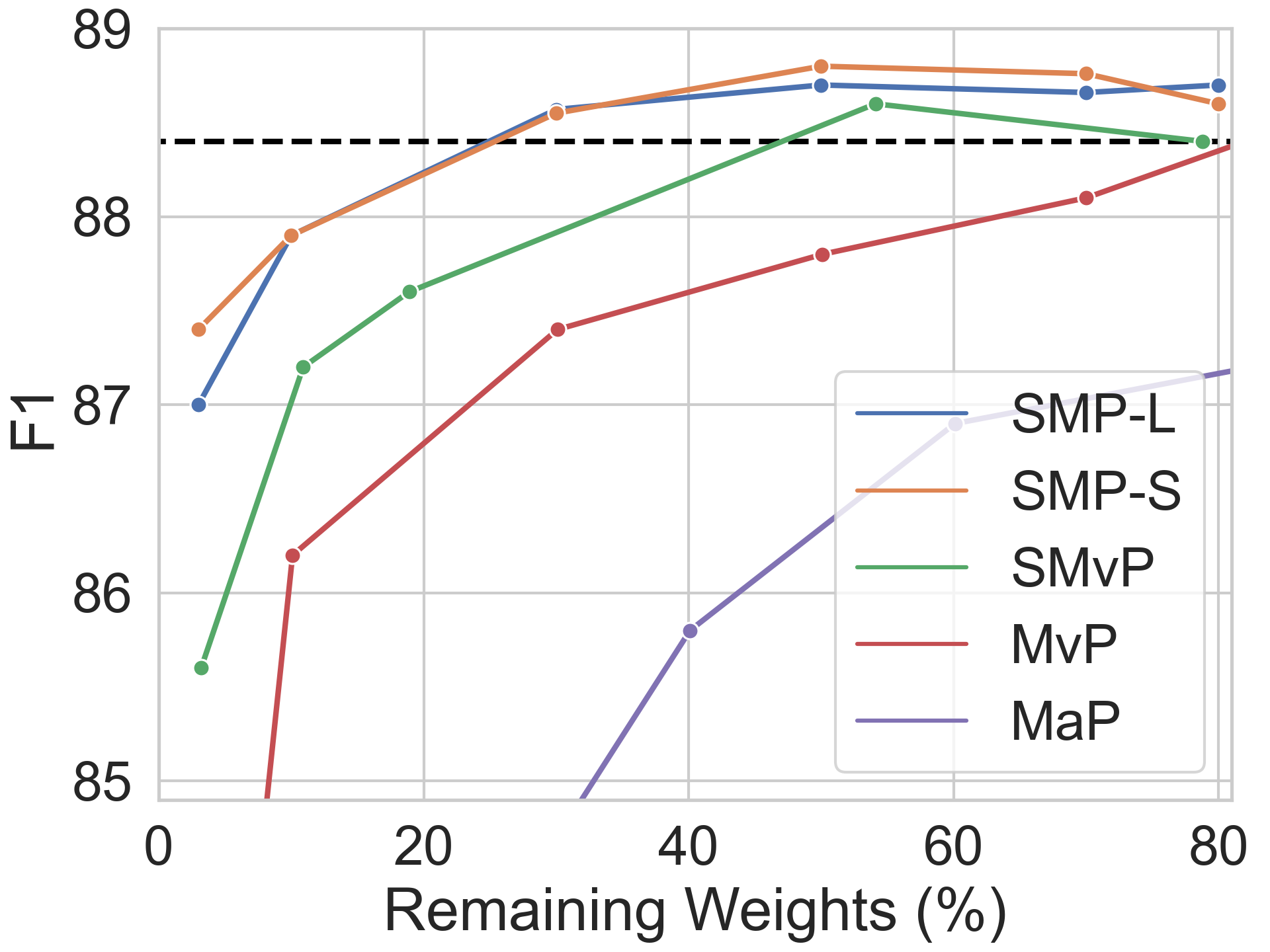}
}
\subfigure[SQuAD + KD]
{
	\includegraphics[width=0.66\columnwidth]{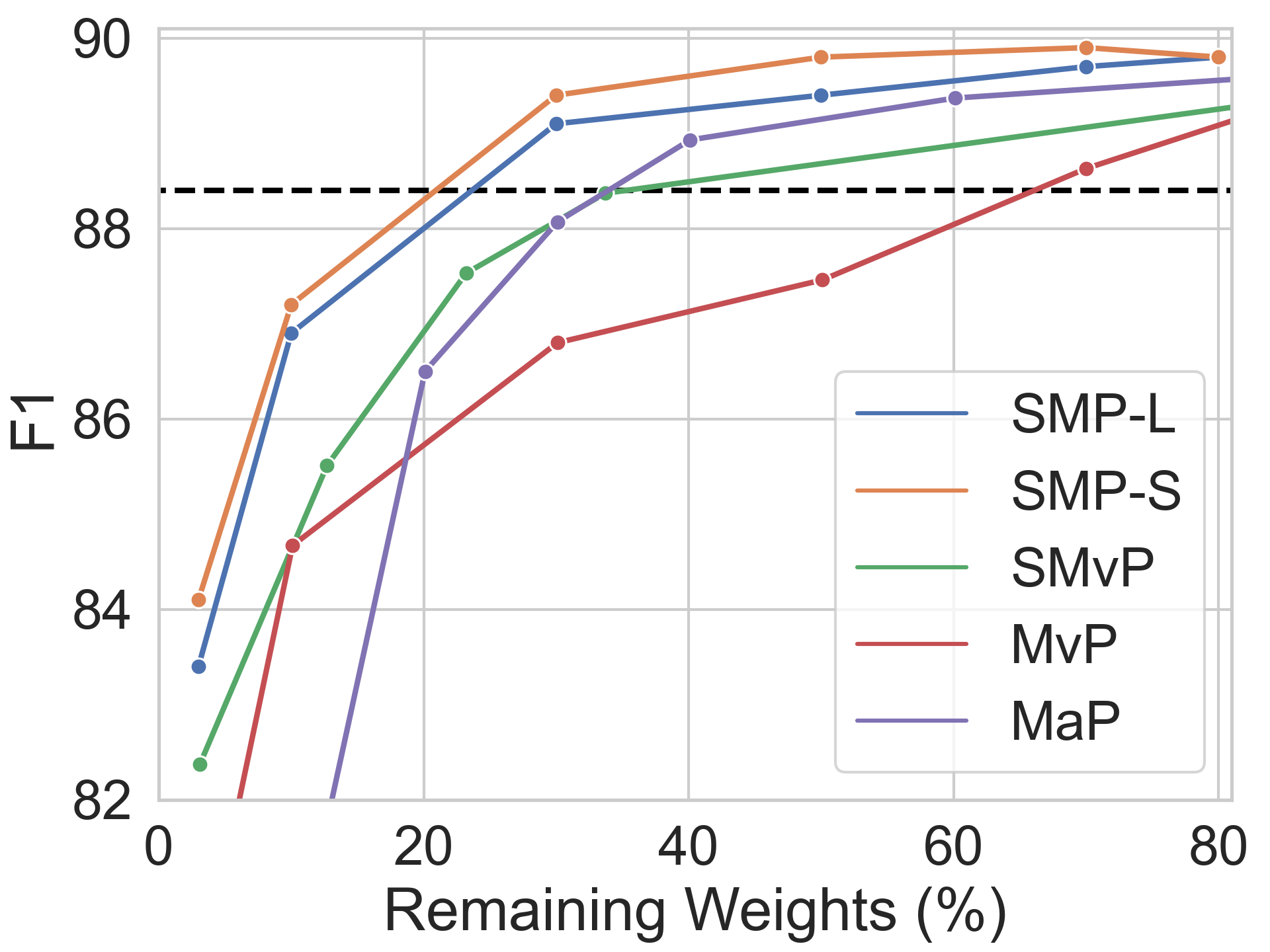}
}

\caption{
  Comparison of different pruning methods from 3\% remaining weights to 80\%.
  The black dashed line in the figures indicates the result of fine-tuned BERT.
  SMvP, MvP and MaP refer to soft-movement pruning, movement pruning and magnitude pruning, respectively.
  KD represents the results with knowledge distillation.
  We report the results of our method on 3\%, 10\%, 30\%, 50\%, 70\%, and 80\% remaining weights.
  Our method constantly outperforms other methods from low sparsity to high.
} %  %大图名称

\label{fig:result}
\end{figure*}

\begin{table*}
  \centering
  \resizebox{0.9\textwidth}{!}{%
   \begin{tabular}{l|cc|ccccccccc}
  \toprule
                & Remaining & New Params & MNLI                      & SST-2                     & MRPC                      & CoLA                       & QNLI                      & QQP                       & RTE                       & STS-B                     &    \\
                & Weights  & Per Task  & $_{\mathrm{M_{ACC}}}$ & $_{\mathrm{ACC}}$ & $_{\mathrm{ACC}}$ & $_{\mathrm{MCC}}$ & $_{\mathrm{ACC}}$ & $_{\mathrm{ACC}}$ & $_{\mathrm{ACC}}$ &  $_{\mathrm{P\ Corr}}$ & Avg.  \\
  \hline
  BERT     & 100\% & 110M       & 84.5                      & 92.9                      & 87.7                      & 58.1                       &  92.0                     & 91.4                      & 71.1                      & 91.2                      & 83.6 \\
  SuperT        & 86.8\% & 98M + $\theta_\textbf{M}$     & 84.5                      & 93.4                      & 86.2                      & 58.8                       & 91.3                      & 91.3                      & 72.5                      & 89.8                      & 83.5 \\
  SMP (Our)           & 80\%     & $\theta_\textbf{M}$   & 85.0                      & 92.9                      & 87.0                      & 61.5                       & 91.5                      & 91.4                      & 72.3                      & 89.6                      & 83.9\\
  \hline
  RoBERTa  & 100\%     & 125M  & 87.6                      & 94.8                      & 90.2                      &63.6                        & 92.8                      & 91.9                      & 78.7                      & 91.2                      & 86.4 \\
  %Our          & 80\%        & 87.6\textsubscript{$\pm$.1} & 94.9\textsubscript{$\pm$.1} & 89.9\textsubscript{$\pm$.6} & 65.4\textsubscript{$\pm$1.1} & 92.8\textsubscript{$\pm$.1} & 91.9\textsubscript{$\pm$.0} & 81.5\textsubscript{$\pm$.3} & 91.1\textsubscript{$\pm$.0} & 86.9 \\
  SMP (Our)           & 80\%   & $\theta_\textbf{M}$    & 87.6                      & 94.9                      & 89.9                      & 65.4                       & 92.8                      & 91.9                      & 81.5                      & 91.1                      & 86.9 \\
  \bottomrule
    \end{tabular}%
    }
  \caption{
    Performance on GLUE development.
    Our results are averaged from five random seeds.
    The results of SuperT are from \cite{Liang2021}, and the remaining weights and new parameters per task in SuperT are averaged over all tasks.
    Note all results are from the setting without knowledge distillation for a fair comparison.
  }
  \label{tab:main-low}
\end{table*}

\section{Experiments}

\subsection{Datasets}
To show the effectiveness of our method, we use three common benchmarks: nature language inference (MNLI)~\cite{N18-1101},
question similarity (QQP)~\cite{aghaebrahimian2017quora} and question answering (SQuAD)~\cite{Rajpurkar2016SQuAD10} following~\citeauthor{Sanh2020}
Moreover, we also use GLUE benchmark~\cite{glue} to validate the performance of our method at low sparsity.
%show the effectiveness of our method in low sparsity regimes with

\subsection{Experiment Setups}
Following previous pruning methods,
we use \texttt{bert-base-uncased} to perform task-specific pruning and report the ratio of remaining weight in the encode.
For the task-specific head, we initial it according to the label words of each task following \cite{gao2020making}.
For SQuAD, we use ``yes'' and ``no'' token embeddings as the weights for starting and ending the classification of answers.
We freeze all weights of BERT including the task-specific head and only fine-tuning mask.
The optimizer is Adam with a learning rate of 2e-2.
The hyperparameter $\lambda_R$ of the regularization term is 400.
We set 12 epochs for MNLI and QQP, and 10 epochs for SQuAD with bath size 64.
For tasks at low sparsity (more than 70\% remaining weights), we set $N$ in cubic sparsity scheduling to 7 epochs.
For tasks at high sparsity, we set $N$ to 3500 steps.
%implement a cubic sparsity scheduling~\cite{cube} with warm-up steps of 0 and pruning steps of 7 epochs included in the overall epochs.

We also report the performance of \texttt{bert-base-uncased} and \texttt{roberta-base} with 80\% remaining weights for all tasks on GLUE with the same batch size and learning rate as above.
For sparsity scheduling, we use the same scheduling for \texttt{bert-base-uncased} and a linear    scheduling for \texttt{roberta-base}.
$N$ in sparsity scheduling is 3500.
For the large tasks: MNLI, QQP, SST2 and QNLI, we use 12 epochs.
For the small tasks: MRPC, RTE, STS-B and COLA, we use 60 epochs.
Note that the above epochs have included pruning steps. For example, we use around 43 epochs to achieve target sparsity in MRPC.
We search the pruning structure from local and global.

\subsection{Baseline}
We compare our method with magnitude pruning~\cite{magnitude}, L$_0$-regularization~\cite{l0}, movement pruning~\cite{Sanh2020} and CAP~\cite{xu2021dense}.
We also compare our method with directly fine-tuning and super tickets~\cite{Liang2021} on GLUE.
For super tickets, it finds that PLMs contain some subnetworks, which can outperform the full model by fine-tuning them.

\subsection{Experimental Results}
Table~\ref{table:main} shows the results of SMP and other pruning methods at high sparsity.
We implement SMP with the local masking function (SMP-L) and our proposed masking function (SMP-S).
%Following previous work, we also validate SMP with and without knowledge distillation.

%RAt 50\% of remaining weights, our method achieves better performance than the original BERT.
%RIn other words, simply removing 50\% parameters of the BERT encoder improves the original BERT performance in MNLI, QQP and SQuAD.
%RMoreover, our method shows non-trivial gains over origin BERT, with improvements over 1.0 for MNLI and SQuAD.
%RThe improvements are more obvious compared to other pruning methods.

SMP-S and SMP-L consistently achieve better performance than other pruning methods without knowledge distillation.
Although movement pruning and SMP-L use the same local masking function,
SMP-L can achieve more than 2.0 improvements on all tasks and sparsity levels in Table~\ref{table:main}.
Moreover, the gains are more significant at 3\% remaining weights.
For soft-movement pruning, which assigns the remaining weights of matrix non-uniformly like SMP-S, it even underperforms SMP-L.

Following previous works, we also report the results with knowledge distillation in Table~\ref{table:main}. %with using baseline BERT$_{\mathrm{base}}$ in Table~\ref{table:main} as teacher.
The improvement brought by knowledge distillation is also evident in SMP-L and SMP-S.
For example, it improves the F1 of SQuAD by 3.3 and 4.1 for SMP-L and SMP-S.
With only 3\% remaining weights, SMP-S even outperforms soft-movement pruning at 10\% in MNLI and QQP.
Compared with CAP, which adds contrastive learning objectives from teacher models,
our method consistently yields significant improvements without auxiliary learning objectives.
For 50\% remaining weights, SMP-S in MNLI achieves 85.7 accuracy compared to 84.5 with full-model fine-tuning, while it keeps all weights of BERT constant.

Our method is also parameter efficient.
Compared with other first-order methods, we can save half of the trainable parameters by keeping the whole BERT and task-specific head frozen.
%Since only importance scores $\textbf{S}$ needs to be fine-tuned, we can save half of the trainable parameters of the other first-order pruning methods.
For new parameters of each task,
it is also an important factor affecting the cost of transferring and storing subnetworks.
Our method only introduces a binary mask $\theta_\textbf{M}$ as new parameters for each task at different sparsity levels, while other methods need to save both $\theta_\textbf{M}$ and the subnetwork.
With remaining weights of 50\%, 10\%, and 3\%, we can save 42.5M, 8.5M, and 2.6M parameters respectively compared with other pruning methods.

\footnotetext{For example at 3\% remaining weights, we can reduce the size of $\theta_\textbf{M}$ to approximately 20\% of its original size through compression.
This means that merely around 0.55M new parameters are introduced at 3\% remaining weights.
Additionally, the compressed $\theta_\textbf{M}$ can be found at  \url{https://github.com/kongds/SMP/releases}.}
Figure \ref{fig:result} shows more results from 3\% remaining weights to 80\% by comparing our method with first-order methods: movement pruning and soft-movement pruning, and the zero-order pruning method: magnitude pruning.
%For CAP, due to it only report the results on
%on three datasets with and without knowledge distillation from high sparsity to low.
We report the results of our method at 3\%, 10\%, 30\%, 50\% and 80\% remaining weights.
Previous first-order methods such as movement pruning underperform magnitude pruning at remaining weights of more than 25\% in MNLI and SQuAD.
Even under high sparsity level like 20\% remaining weights, magnitude pruning still strongly outperforms both movement pruning and soft-movement pruning in Figure \ref{fig:result} (c).
This shows the limitation of current first-order methods that performing ideally only at very high sparsity compared to zero-order pruning methods.
However, SMP-L and SMP-S as first-order methods can constantly show better performance than magnitude pruning at low sparsity.
For the results without knowledge distillation, SMP-S and SMP-L achieve similar performance of soft-movement pruning with much less remaining weights.
Considering to previous LTH in BERT,
we find SMP-S can outperform full-model fine-tuning at a certain ratio of remaining weights in Figure \ref{fig:result} (a), (b) and (c), indicating that BERT contains some subnetworks that outperform the original performances without fine-tuning.
For the results with knowledge distillation,
SMP-S and SMP-L benefit from knowledge distillation at all sparsity levels.
After removing even 70\% weights from the encoder, our method still strongly outperforms full-model fine-tuning.

We also validate our method on GLUE and report the results at 80\% remaining weights in Table~\ref{tab:main-low}.
Compared to full-model fine-tuning, our method achieves better performance on two PLMs by only removing 20\% parameters in the encoder while keeping the remaining parameters unchanged.
Compared to SuperT, which searches 8 different sparsity levels for each task,
our method achieves better performance by using the same sparsity levels.
In addition, our method also saves more than 98M new parameters per task compared to SuperT.
%We also show results at low sparsity levels in Figure~\ref{fig:result}.
\begin{table}[t]
\centering
\scalebox{0.8}{
    \begin{tabular}{ccccccccc}
    \toprule
    & {Masking} & \multicolumn{3}{c}{ MNLI} &
    \multicolumn{3}{c}{ SQuAD} \\
    \cmidrule(lr){3-5} \cmidrule(lr){6-8}
   & {Function} & 80\% & 10\% & 3\% & 80\% & 10\% & 3\% \\
    \midrule
    $T$ & $\sigma(\mathbf{S_{(\cdot)}}^l) > \tau $ & N/A & N/A & N/A & N/A & N/A & N/A \\
    $G$ & $\mathbf{S_{(\cdot)}}^l \geq S^{v} $ & 85.0 & 81.0 & 80.1 & 88.2 & 83.1 & 79.3 \\
    $L$ & $\text{Top}_v(\mathbf{S_{(\cdot)}}^l)$  & 84.8 & 82.0 & 80.6 & 88.0&  84.3 & 81.0 \\
    $S$ & $\text{Top}_{v_{(\cdot)}^l}(\mathbf{S_{(\cdot)}}^l)$ & 85.0 & 82.5 & 80.9 & 88.3& 84.6 & 81.4 \\
    \bottomrule
    \end{tabular}
}
\caption{
  Influence of different masking functions.
  We report the results in MNLI and SQuAD with 80\%, 10\% and 3\% remaining weights.
  N/A means that our method with corresponding masking function fails to converge in our setting.
  Masking function is to transform $\mathbf{S}_{(\cdot)}$ to the binary mask $\mathbf{M}_{(\cdot)}^l$ of   $\mathbf{W}_{(\cdot)}^l$.
  $T$ refers to the thresholding masking function following~\cite{Sanh2020}, and $\tau$ is the threshold.
  $G$ and $L$ are global and local masking functions, and $S^{v}$ is the smallest value in the top $v$\% after sorting all $\mathbf{S}$ together.
  $S$ refers to our proposed masking function, and $v_{(\cdot)}^l$ is from Eq.~\ref{eq:our_v}.
}
\label{table:idmf}
\end{table}

\section{Analysis}

\begin{figure}[h]
\centering    %居中
\subfigure[$\mathbf{W}_Q$]
{
	\includegraphics[width=0.45\columnwidth]{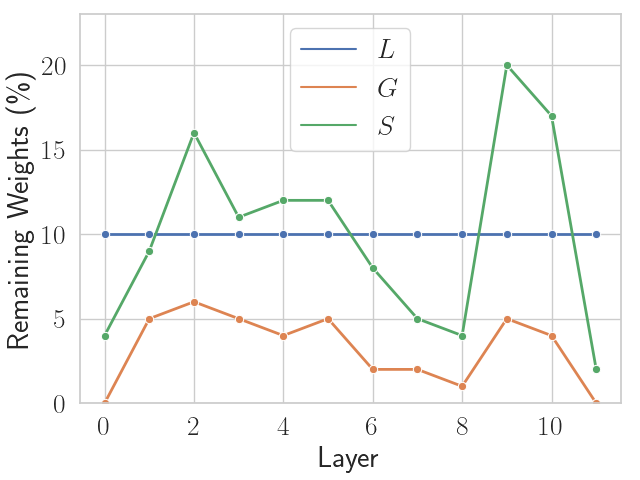}
}
\subfigure[$\mathbf{W}_K$]
{
	\includegraphics[width=0.45\columnwidth]{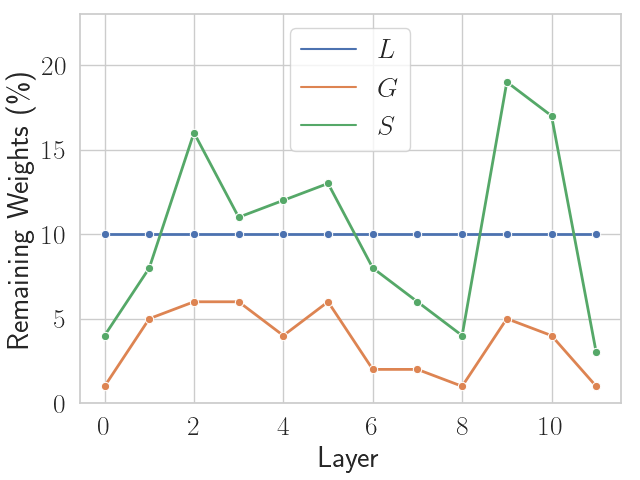}
}
\subfigure[$\mathbf{W}_V$]
{
	\includegraphics[width=0.45\columnwidth]{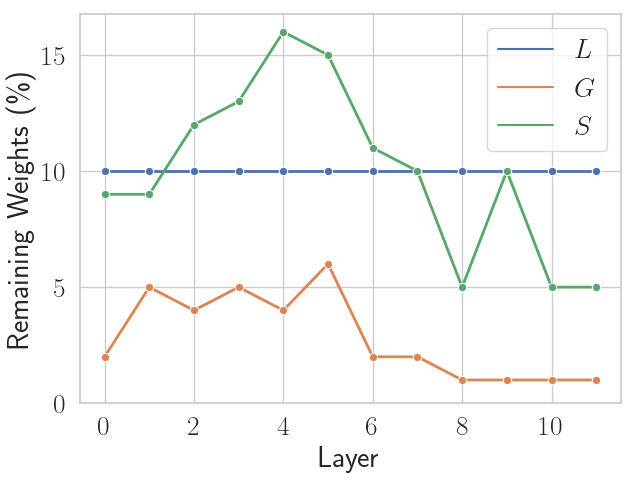}
}
\subfigure[$\mathbf{W}_O$]
{
	\includegraphics[width=0.45\columnwidth]{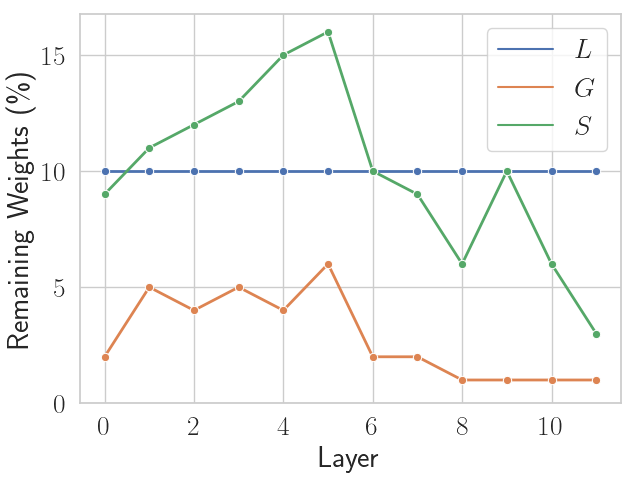}
}
\subfigure[$\mathbf{W}_U$]
{
	\includegraphics[width=0.45\columnwidth]{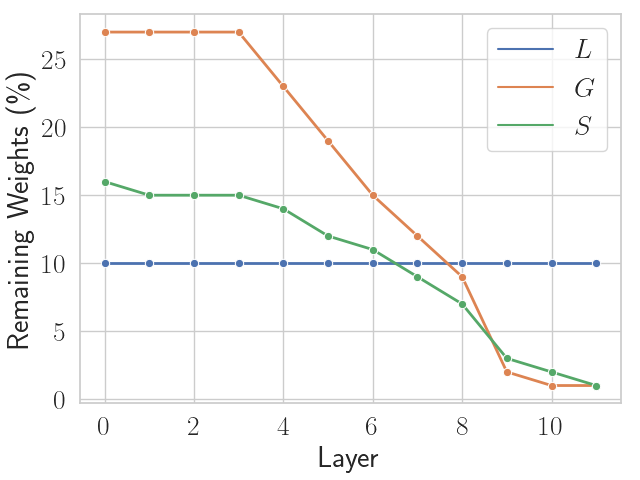}
}
\subfigure[$\mathbf{W}_D$]
{
	\includegraphics[width=0.45\columnwidth]{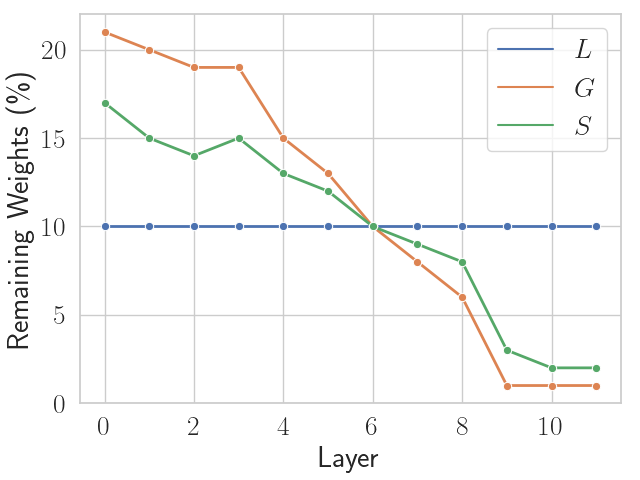}
}
\subfigure[Overall]
{
	\includegraphics[width=0.95\columnwidth]{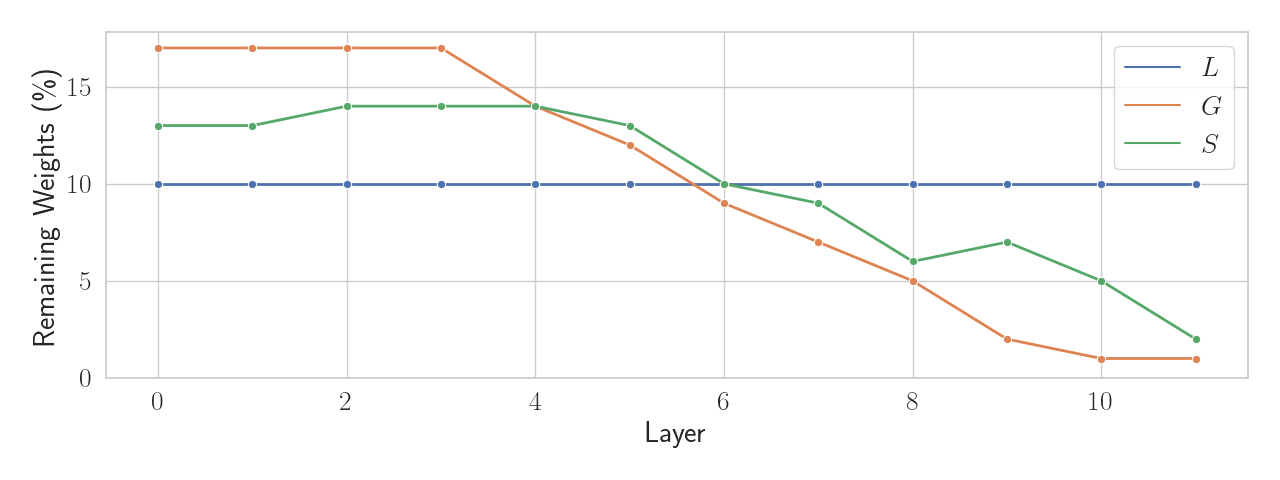}
}
\caption{
%The x axis is epoch and the y axis is training loss.
Distribution of remaining weights corresponding to each layer.
Overall refers to the overall remaining weights of each layer.
$\mathbf{W}_{(\cdot)}$ is the remaining weights for each weight matrix in BERT encoder.
$L$, $G$ and $S$ in figures refer to the masking functions following Table~\ref{table:idmf}.
}
\label{fig:dis}
\end{figure}

\begin{figure}[t]
\centering
% \vspace{-0.1in}
\includegraphics[width=\linewidth]{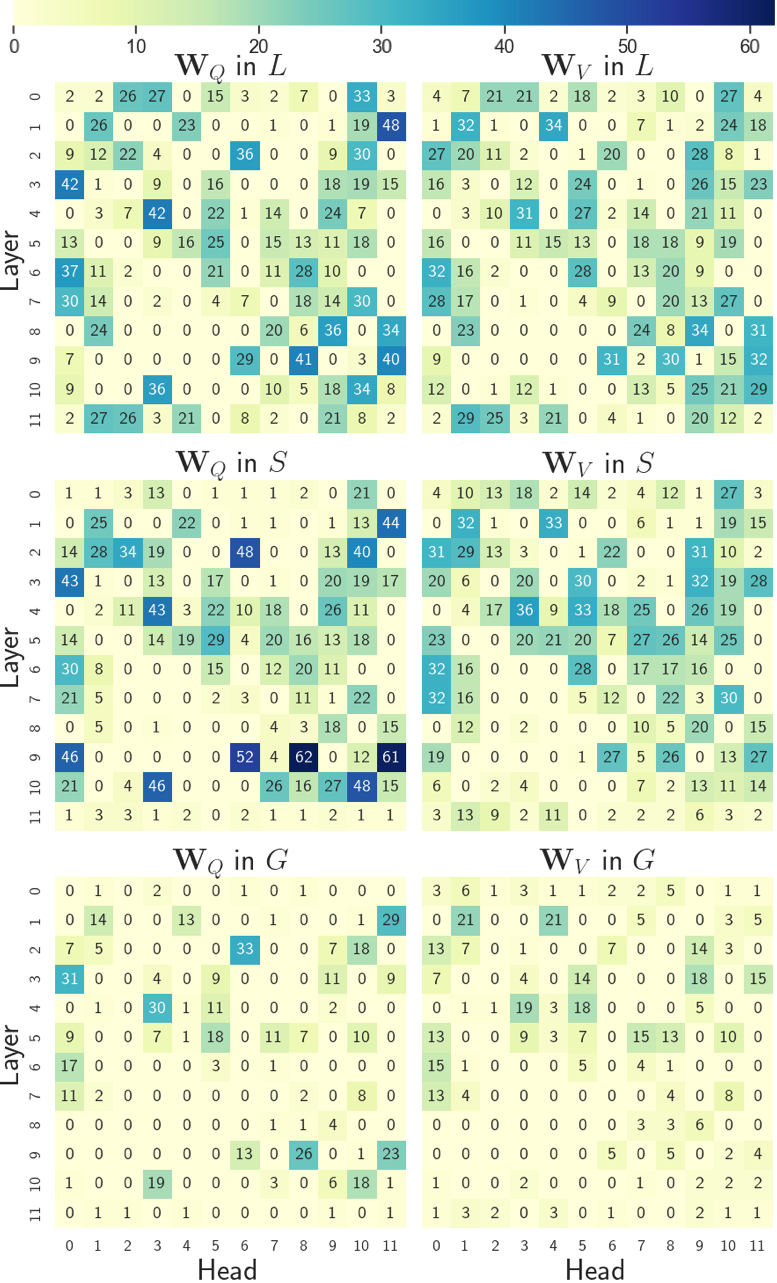}
\caption{
  Remaining weights ratio per attention head of $\mathbf{W}_Q$ and $\mathbf{W}_V$ in MNLI with 10\% remaining weights.
  %We omit $W_K$ due to it has a similar distribution as $W_Q$.
  Each cell refers to the remaining weights ratio of the corresponding attention head.
  The darker the color, the higher the ratio of remaining weight.
  $L$, $G$ and $S$ in figures refer to the masking functions following Table~\ref{table:idmf}.
}
\label{fig:head}
\end{figure}

\subsection{Masking Function} \label{sec:masking}
In this section, we discuss the influence of different masking functions.
Table \ref{table:idmf} shows the results of different masking functions on our method without knowledge distillation.
Contrary to previous pruning methods, the thresholding masking function $T$ fails to converge in our method due to the difficulty in controlling the sparsity during training.
For global masking function $G$,
we sort all 85M BERT encoder weights and remain Top $v$\% weights in each training step.
Compared to local masking functions $L$, $G$ takes more than twice the training times due to the computational cost of sorting 85M weights.
Although it took the longest to train, it still underperforms $L$ at 10\% and 3\% remaining weights.
Contrary to $G$,
our proposed masking function $S$ outperforms $L$ without additional training time since $S$ directly assigns the remaining weights of each matrix.
More results of masking functions $S$ and $L$ are also available in Table~\ref{table:main} and Figure~\ref{fig:result}.

Figure \ref{fig:dis} displays the distribution of remaining weights in different layers in MNLI with 10\% remaining weights.
We find \textit{G} assigns too many remaining weights for $\mathbf{W}_U$ and $\mathbf{W}_V$, which are four times larger than other matrices.
It causes other weight matrices such as $\mathbf{W}_Q$ to be more sparse than \textit{S} and \textit{L}.
Following previous studies~\cite{Sanh2020, Mallya2018PiggybackAM},
we also find that overall sparsity tends to increase with the depth of the layer.
However, only $\mathbf{W}_U$ and $\mathbf{W}_V$ follow this pattern in all three matrices.
Since $\mathbf{W}_U$ and $\mathbf{W}_V$ occupy more than 60\% of the weight in each layer,
it causes the overall distribution of each layer also follows their trend as well.
%The sparsity of remaining matrices is not following this pattern on $L$ and $G$.
%For matrices in attention heads, they do not follow this trend.

To understand the behavior of attention heads,
we also display the remaining weights ratio of each head in Figure~\ref{fig:head}.
Each row represents a matrix containing 12 heads.
Due to space limitation and the similar distribution between $\mathbf{W}_Q$ and $\mathbf{W}_K$, we only show $\mathbf{W}_Q$ and $\mathbf{W}_V$.
%we find that $\mathbf{W}_Q$ and $\mathbf{W}_K$ have a similar distribution of attention heads in all three masking functions.
%So we only  for space limitation.
Instead of assigning sparsity uniformly to each head,
the sparsity of each head is not uniform in three masking functions, with most heads having only below 1\% or below remaining weights.
Furthermore, three masking functions show similar patterns even with different ways of assigning remaining weights.
For our masking function $S$,
$S$ can assign more remaining weights to important heads compared to $L$, and some heads in $\mathbf{W}_Q$ achieve more than 60\% remaining weights at 9th layer.
For global masking function $G$,
due to most of remaining weights being assigned to $\mathbf{W}_U$ and $\mathbf{W}_D$,
%attention heads in $G$ are more sparsity than $S$ and $L$.
the average remaining weights ratio of $\mathbf{W}_Q$ and $\mathbf{W}_V$ in $G$ are only 3.2\% and 2.8\%, which causes $G$ to underperform other masking functions.
%Under these sparsity levels, most heads are masked with 0\% remaining weights.

\subsection{Task-Specific Head}
To validate the effectiveness of our task-specific head initialization method, we compare it with training from scratch.
\label{table:tsh}
\begin{table}[H]
\centering
\scalebox{0.8}{
    \begin{tabular}{ccccccc}
    \toprule
     & \multicolumn{3}{c}{ MNLI} &
    \multicolumn{3}{c}{ SQuAD}  \\
    \cmidrule(lr){2-4} \cmidrule(lr){5-7}  &
    80\% &10\% & 3\% &80\% & 10\% & 3\%  \\
    \midrule
 From scratch& 84.6 & 81.7& 80.5 & 87.5 & 84.2 & 80.7   \\
 Initialization   & 84.8 & 82.0 & 80.6 & 88.0 & 84.3 & 81.0  \\
    \bottomrule
    \end{tabular}
}
\caption{
  Influence of different task-specific head methods.
  ``From scratch'' refers to training head from scratch following previous pruning methods.
  ``Initialization'' refers to our initialization method.
}
\label{table:tsh}
\end{table}
Table~\ref{table:tsh} shows the results of SMP-L in MNLI and SQuAD with 80\%, 10\% and 3\% remaining weights.
For training from scratch, we randomly initial the head and fine-tune it with the learning rate of 3e-5 following previous pruning methods.
Results show our method achieves better performance with task-specific heads frozen.

\subsection{Training Objective}
Regularization term in training objective is a key factor for our method.
We find that our method is hard to converge at high sparsity without regularization term $R$ in Table~\ref{table:rsh}.
With the increase of sparsity, the performance gap between with and without $R$ sharply increases.
SMP-L without $R$ even fails to converge at 10\% and 3\% remaining weights in SQuAD.

%MHA and FFN are forumulated as:
%\begin{equation}
%\operatorname{MHA}(X)=\sum_{i=1}^{N_{h}} \operatorname{Att}\left(W_{Q}^{(i)}, W_{K}^{(i)}, W_{V}^{(i)}, W_{O}^{(i)}, X\right)
%\operatorname{FFN}(X)=\operatorname{gelu}\left(X W_{U}\right) \cdot W_{D}
%\end{equation}
\begin{table}[h]
\centering
\scalebox{0.8}{
    \begin{tabular}{lcccccc}
    \toprule
     & \multicolumn{3}{c}{ MNLI} &
    \multicolumn{3}{c}{ SQuAD}  \\
    \cmidrule(lr){2-4} \cmidrule(lr){5-7}  &
    80\% &10\% & 3\% &80\% & 10\% & 3\%  \\
    \midrule
 SMP-L   & 84.8 & 82.0 & 80.6 & 88.0 & 84.3 & 81.0  \\
 \ \ \ w/o $R$ & 84.2 & 80.1 & 69.2 & 86.6 & N/A & N/A \\
    \bottomrule
    \end{tabular}
}
\caption{
  Influence of regularization term.
  $R$ refers to the regularization term.
  N/A refers to unable convergence.
}
\label{table:rsh}
\end{table}

As analyzed in section~\ref{sec:to}, %$R$ encourages $\mathbf{M}$ to be changed % that $R$ encourage masking change,
we find the remaining weights in attention heads are more uniform without $R$.
For example, the standard deviation of remaining weights in each attention head is 3.75 compared to 12.4 in SMP-L with $R$ in MNLI with 10\% remaining weights.
In other words, without $R$, it cannot assign more remaining weights to important heads as in Figure~\ref{fig:head}.

\section{Conclusion}
In this paper, we propose a simple but effective task-specific pruning method called Static Model Pruning (SMP).
Considering previous methods, which perform both pruning and fine-tuning to adapt PLMs to downstream tasks, we find fine-tuning can be redundant since first-order pruning already converges PLMs.
Based on this, our method focuses on using first-order pruning to replace fine-tuning.
Without fine-tuning,
our method strongly outperforms other first-order methods.
Extensive experiments also show that our method achieves state-of-the-art performances at various sparsity.
For the lottery ticket hypothesis in BERT, we find it contains sparsity subnetworks that achieve original performance without training them, and these subnetworks at 80\% remaining weights even outperform fine-tuned BERT on GLUE.

\section{Limitation}
Like all unstructured pruning methods, SMP is hard to achieve inference speedup compared to structured pruning methods. Since SMP prunes model without fine-tuning, this also limits the extension of SMP to structured pruning methods. However, we find that most rows of the sparsity matrices in SMP are completely pruned at high sparsity level. This allows us to directly compress the size of matrices, resulting in faster inference. For example, the 3\% remaining weights model of MNLI can be compressed to 47.43\% of the model actual size (resulting in around 1.37$\times$ inference speedup) without retraining or performance loss. By removing rows of matrices that contain less than 10 remaining weights, we can further compress it to 25.19\% actual size (1.76$\times$ inference speedup) with 0.9 accuracy drop. We expect that a carefully designed loss function during training could result in even smaller actual model size and faster inference speedup, which we leave it in the future.

\section{Acknowledgments}
The research work is supported by the National Key Research and Development Program of China under Grant No. 2021ZD0113602, the National Natural Science Foundation of China under Grant Nos. 62276015, 62176014, the Fundamental Research Funds for the Central Universities.

\bibliography{anthology,custom}

\appendix
\section{Standard Deviation of Tasks}
We also report our standard deviation of tasks from 5 random runs in Table \ref{table:std1} and \ref{table:std2}.

\begin{table}[h]
\centering
\scalebox{0.8}{
    \begin{tabular}{ccccccc}
      \hline
     & &\multicolumn{3}{c}{with KD} & \multicolumn{2}{c}{without KD} \\
    \cmidrule(lr){3-5} \cmidrule(lr){6-7}
    & &50\% & 10\% & 3\% & 10\% & 3\% \\
    \hline
    MNLI & SMP-L & 0.17 & 0.26 & 0.19 &  0.27 & 0.20 \\
    $\mathrm{M_{ACC}\ std.}$   & SMP-S & 0.13 & 0.24 & 0.30 & 0.25 & 0.28\\
    \hline
    QQP & SMP-L & 0.04 & 0.01 & 0.08 & 0.06 & 0.01\\
    $\mathrm{ACC \ std.}$   & SMP-S & 0.02 & 0.03 & 0.02 & 0.01 & 0.02 \\
    \hline
    SQuAD   & SMP-L & 0.17 & 0.09 & 0.03 &  0.36 & 0.01 \\
    $\mathrm{F1\ std.}$   & SMP-S & 0.10 & 0.07 & 0.02 & 0.42 & 0.07\\
    \hline
    \end{tabular}
}
\caption{
  Standard deviation of Table \ref{table:main}
}
\label{table:std1}
\end{table}

\begin{table}[h]
\centering
\scalebox{0.8}{
    \begin{tabular}{ccc}
      \hline
      & SMP(BERT) & SMP(RoBERTa) \\
      \hline
MNLI&0.15  & 0.12 \\
QNLI&0.15  & 0.11 \\
QQP &0.03  & 0.14 \\
SST2&0.36  & 0.28 \\
MRPC& 1.21 & 0.44 \\
COLA&0.69  & 0.65 \\
STSB&0.14  & 0.16 \\
RTE &1.59  & 0.74 \\
      \hline
    \end{tabular}

}
\caption{
  Standard deviation of Table \ref{tab:main-low}
}
\label{table:std2}
\end{table}

\end{document}